\documentclass{article}

\usepackage[nonatbib, preprint]{neurips_2019}

\usepackage[utf8]{inputenc} % allow utf-8 input
\usepackage[T1]{fontenc}    % use 8-bit T1 fonts
\usepackage{hyperref}       % hyperlinks
\usepackage{url}            % simple URL typesetting
\usepackage{booktabs}       % professional-quality tables
\usepackage{amsfonts}       % blackboard math symbols
\usepackage{nicefrac}       % compact symbols for 1/2, etc.
\usepackage{microtype}      % microtypography

\usepackage{graphicx}
\usepackage{amsmath}
\usepackage{amssymb}
\usepackage{xcolor}
\usepackage{tabularx}
\usepackage{multirow}
\usepackage{colortbl}

\definecolor{Gray}{gray}{0.86}
\newcolumntype{a}{>{\columncolor{Gray}}c}

\title{Continual Reinforcement Learning in 3D Non-stationary Environments}

\author{%
  Vincenzo Lomonaco\\
  University of Bologna\\
  Bologna, Italy \\
  \texttt{vincenzo.lomonaco@unibo.it} \\
  \And
  Karan Desai \\
  University of Michigan \\
  Ann Arbor, United States \\
  \texttt{kdexd@umich.edu} \\
  \AND
  Eugenio Culurciello \\
  Purdue University \\
  West Lafayette, United States \\
  \texttt{eugenio.culurciello@purdue.org} \\
  \And
  Davide Maltoni \\
  University of Bologna\\
  Bologna, Italy \\
  \texttt{davide.maltoni@unibo.it} \\
}

\begin{document}

\maketitle

\begin{abstract}
 High-dimensional always-changing environments constitute a hard challenge for current reinforcement learning techniques. Artificial agents, nowadays, are often trained off-line in very static and controlled conditions in simulation such that training observations can be thought as sampled i.i.d. from the entire observations space. However, in real world settings, the environment is often non-stationary and subject to unpredictable, frequent changes. In this paper we propose and openly release \textit{CRLMaze}, a new benchmark for learning continually through reinforcement in a complex 3D non-stationary task based on ViZDoom and subject to several environmental changes. Then, we introduce an end-to-end model-free continual reinforcement learning strategy showing competitive results with respect to four different baselines and not requiring any access to additional supervised signals, previously encountered environmental conditions or observations.
\end{abstract}

\section{Introduction}

In the last decade we have witnessed a renewed interest and major progresses in reinforcement learning (RL) especially due to recent deep learning developments \cite{Arulkumaran2017}. State-of-the-art RL agents are now able to tackle fairly complex problems involving high-dimensional perceptual data, which were even unthinkable to solve without explicit supervision before \cite{Mnih2015, Silver2017}.

However, much of these progresses have been made in very narrow and isolated tasks, often in simulation with thousands of trials and with the common assumption of a stationary, fully-explorable environment from which to sample observations i.i.d. or approximately so. Even in the case of more complex tasks and large environments, a common technique known as \emph{memory replay} \cite{Mnih2015, Isele2018, Liu2019b} is adopted, consisting in storing old observations in an external memory buffer to simulate an i.i.d. sampling. Roughly the same result can be also achieved through multiple replicas of the agent randomly spawned in the environment and collecting several different observations at the same time, hence approximating the coverage of the entire observations space \cite{Mnih2016}.

Nevertheless, dealing with single agents in the real-world and subject to computational and memory constraints these solutions suddenly appear less practical. This is especially true with \emph{always-changing} environments and multi-task settings where re-sampling is impossible and storing old observations is no longer an option since it would require a constant grow in terms of memory consumption and computational power needed to re-process these observations. On the other hand, if the memory replay buffer is limited in size, the agent suddenly incurs in the phenomenon known in literature as \emph{catastrophic forgetting}, being unable to retain past knowledge and skills in previously encountered environmental conditions or tasks \cite{McCloskey1989, Robins1995, French1999, lomonaco2018thesis}.

Learning continually from data is a topic of steadily growing interest for the machine learning community and concerns itself with the idea of improving adaptation and generalization capabilities of current machine learning models by providing efficient updating strategies when new observations become available without \emph{storing}, \emph{re-sampling} or \emph{re-processing} the previous ones (or as little as possible). While much of the focus and research efforts in continual learning have been devoted to multi-task settings (where a single model is exposed to a sequence of distinct and well-defined tasks over time) \cite{Parisi2019, Chen2018}, several practical scenarios would also benefit artificial agents that learn continually in complex non-stationary reinforcement environments.

% explain that we don't use t and other stuff
In this paper, we focus on the more complex problem of a single task, constantly changing over time. As it has been shown in some supervised contexts, the clear separation in tasks (i.i.d. by parts), along with the presence of a supervised \emph{``task label'} $t$ \cite{Diaz-Rodriguez2018}, greatly helps taming the problem of forgetting \cite{maltoni2019, Aljundi2018}. We argue that learning without any notion of task or distributional shift (both during training and inference), at least from an external oracle, is a more natural approach worth pursuing for improving the autonomy of every artificial learning agent.

The original contributions of this paper can be summarized as follows:

\begin{itemize}
   %\item We discuss and argue about the importance on the focus on non-stationary environments rather than explicit and supervised multi-task settings for autonomous reinforcement learning agents.
   \item We design and openly release a new benchmark, \emph{CRLMaze} based on VizDoom \cite{Kempka2016}, for assessing continual reinforcement learning (CRL) techniques in an always-changing object-picking task. \emph{CRLMaze} is composed of 4 scenarios (\emph{Light}, \emph{Texture}, \emph{Object}, \emph{All}) of incremental difficulty and a total of 12 maps. To the best of our knowledge, this is one of the first attempts to scale continual reinforcement learning to complex 3D non-stationary environments.
   \item We provide 4 continual reinforcement learning baselines for each scenario. 
   \item We propose an end-to-end, model-free continual reinforcement learning strategy, \emph{CRL-Unsup}, which is agnostic to the changes in the environment and does not exploit a \emph{memory replay} buffer or any distribution-specific \emph{over-parametrization}, showing competitive results with respect to the supervised baselines (see section \ref{sec:exp}). The core insight of our strategy is to consolidate past memories through regularization as in \cite{Kirkpatrick2017}, but proportionally to the difference between the expected reward and the actual reward (hence encoding a novel environmental condition in which the agent is unable to operate).
\end{itemize}

\begin{figure}[t]
  \centering
  \includegraphics[width=0.80\textwidth]{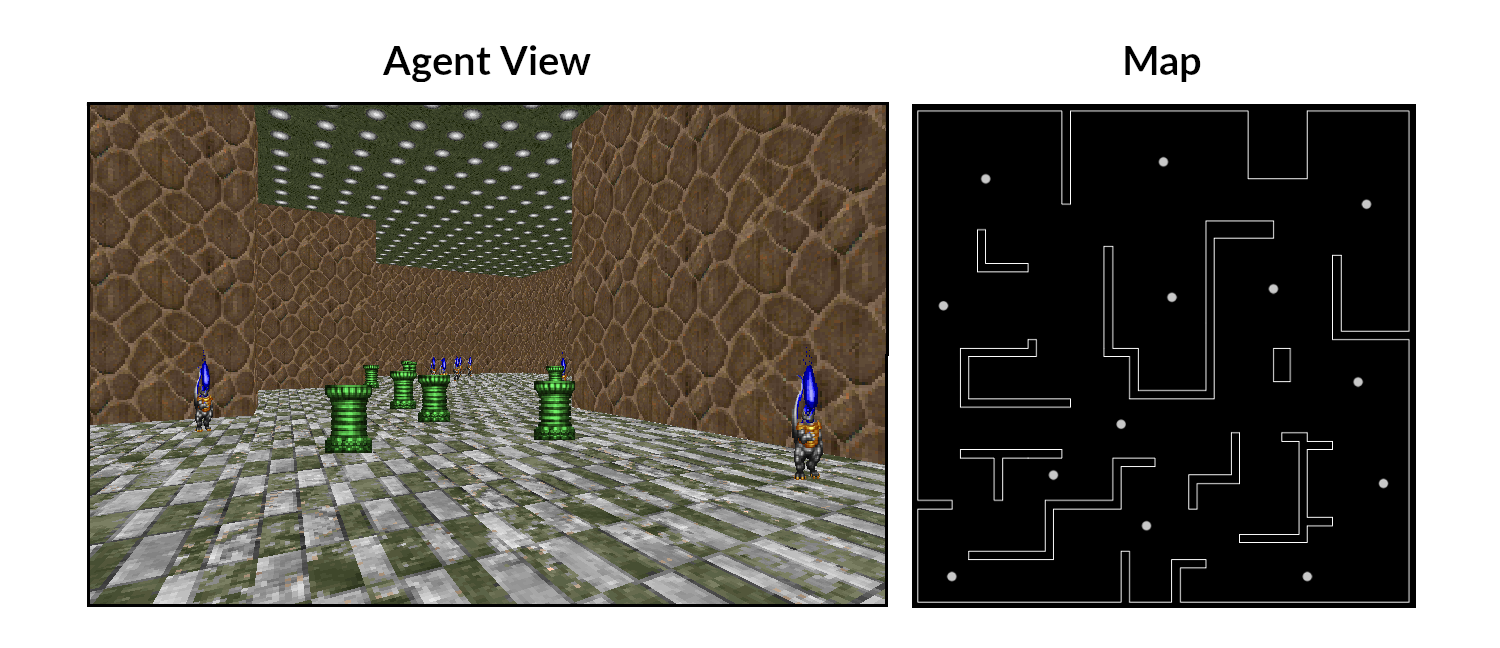}
  \caption{\small The 3D maze environment developed with \emph{ZDoom} and \emph{Slade3}. On the left, an example image from the point of view of the agent is reported. On the right, the planar view of the maze structure is shown. White points on the map represent random spawning points used by the agent during both training and test episodes. Better viewed in colors.}
  \label{img:doom-maze}
\end{figure}

% sections intro
All the environments and the code to reproduce and expand the experiments discussed in this paper are available at: \url{https://github.com/vlomonaco/crlmaze}. 

The rest of the paper is organized as follows: in Section \ref{sec:crlmaze}, the \emph{CRLMaze} benchmark is described; In Section \ref{sec:strat}, the CRL strategies used for the experiments reported in Section \ref{sec:exp} are outlined. Finally, in Section \ref{sec:conclusions}, key questions and future work in this area are discussed.

\section{CRLMaze: a 3D Non-stationary Environment}
\label{sec:crlmaze}

Continual Learning (CL) in reinforcement learning environments is still in its infancy. Despite the the obvious interest in applying CL to less supervised settings and the early, promising results in this context \cite{Ring94,thrun1995lifelong}, reinforcement learning tasks constitute a much more complex challenge where it is generally more difficult to disentangle the complexity introduced by distributional shifts from those introduced by the lack of a strong supervision. 

It is also worth noting that state-of-the-art reinforcement learning algorithms and current hardware computational capabilities does not make experimentations and prototyping easily accomplished on complex environments where physical simulation constitute an heavy computational task per se. In a continual learning context, the problem becomes even harder since an exposition of the same model to sequential streams of observations is needed (and cannot be parallelized by definition). This is why recent reinforcement learning algorithms for continual learning have been tested only on arguably simple tasks of low/medium input space dimension and complexity \cite{Kirkpatrick2017, Al-shedivat2017,Nagabandi2018}.

% Plaforms and softwares for RL envs.
Nevertheless, at the same time, state-of-the-art reinforcement learning algorithms have started tackling more complex problems in 3D static environments. \emph{VizDoom} \cite{Kempka2016}, followed soon after by other research platforms like \emph{DeepMind Labs} \cite{Beattie2016} and \emph{Malmo} \cite{Johnson2016-malmo}, allowed researchers to start exploring new interesting research directions with the aim of scaling up current reinforcement learning algorithms. 

\begin{figure}[t]
  \centering
  \includegraphics[width=0.90\textwidth]{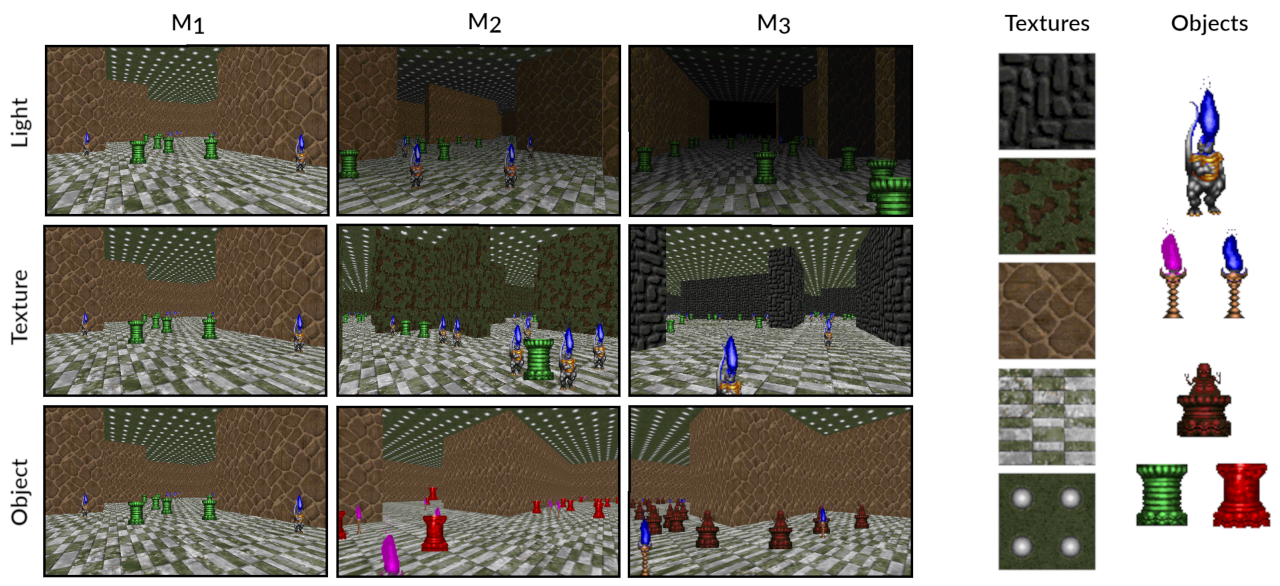}
  \caption{\small On the left, the environmental changes for each of scenario (\emph{Light}, \emph{Texture}, \emph{Object}) in the 3D \emph{CRLMaze}. For the \emph{All} scenario, each map is the composition of all the environment variations introduced in the respective maps of all the other scenarios. In all the cases, changes are not gradual but occur abruptly at three (equidistant) points in time. On the right, textures and objects used for the \emph{CRLMaze} scenarios. Better viewed in colors.}
  \label{img:env-changes}
\end{figure}

% Intro vizdoom
\emph{VizDoom} is a reinforcement learning API build around the famous \emph{ZDoom} game engine and providing all the necessary utilities to train a RL agent in arbitrary complex environments. This framework is particularly interesting since it has been open-sourced to both Windows and Unix systems and it was already built on the idea of flexibility and customizability, allowing users to create custom maps and modify behaviorial responses of the environment through the simple \emph{Action Code Script} (ACS) language. 

% intro Vizdoom task
In this paper, we propose an original 3D ViZDoom environment for continual reinforcement learning and an object-picking task named \textbf{\emph{CRLMaze}}\footnote{In particular, we used \emph{Slade3} as the environment editor.} (see Fig. \ref{img:doom-maze}). The task consists of learning how to navigate in a complex maze and pick up \emph{``column bricks''} while avoiding \emph{``flaming lanterns''} (see Fig. \ref{img:env-changes}). However, the environment in this case is \emph{non-stationary} meaning that is subject to several environmental changes leading to major difficulties for standard reinforcement learning algorithms.

% descripion of light, texture, objects, all
For properly assessing novel continual reinforcement learning strategies in the aforementioned 3D complex environment we split the benchmark in four different scenarios of incremental difficulty with respect to different environmental changes (see Fig. \ref{img:env-changes}):

\begin{itemize}
  \item \emph{\textbf{Light}}: In this scenario the illumination of the environment is altered over time. While intuitively this scenario may appear as one of the easiest, as we will see in the experimental section \ref{sec:exp}, it constitutes one of the most difficult since visual features from the environment do not change only in terms of RGB pixel magnitudes by a scalar factor, but also in terms of agent visibility (i.e. the radius in the 3D space up to which the RGB colors saturates to complete black), as shown in Fig. \ref{img:env-changes} (top row).
  \item \emph{\textbf{Texture}}: In this scenario walls textures are changed over time. The ability to pass over invariant features of the background is often taken for granted in many supervised tasks with state-of-the-art deep architectures \cite{Peng15}. However, as shown in the past, reinforcement learning agents are quite fragile also with respect to minor environmental changes \cite{Kansky17}.
  \item \emph{\textbf{Object}}: In this scenario the shapes and colors of the objects are changed over time. Invariance with respect to object shapes and colors is another important property every learning system should possess when facing real-world conditions where surrounding objects appearances are subject to constant changes due to deterioration and substitutions.
  \item \emph{\textbf{All}}: In this environment lights, textures as well as objects are subject to change over time. This scenario is also proposed with the idea of providing a comprehensive scenario for 3D environments in complex non-stationary settings, combining all the environmental condition variations proposed in the previous ones.
\end{itemize}

\begin{table*}[!htbp]
%\small
\centering
 \caption{\small Some common environments used for continual or meta reinforcement learning. The proposed benchmark, CRLMaze (bottom), shows significant advancements in terms of task complexity and non-stationary elements.}
 \label{tab:envs}
\vspace{5px}
\setlength\tabcolsep{2.5pt}
\small
\begin{tabular}{l|c|c|p{6.5cm}}
\toprule
\textbf{Environment Name} & \textbf{Input Dim.} & \textbf{3D} & \textbf{Non-Stationary Elements}\\
\cmidrule{1-4}
Locomotion Environment \cite{Al-shedivat2017} & 14 & yes & 2 over 16 joints torques are scaled down by a constant factor.\\
MiniGrid \cite{Chevalier2019} & 6$\times$6$\times$3 & no & ``Competencies'' introduced in a curriculum.\\
Catcher \cite{Riemer2018} & 256$\times$256$\times$3 & no & Vertical velocity of pellet increased of 0.03 from default 0.608.\\
Flappy Bird \cite{Riemer2018} & 288$\times$512$\times$3 & no & Pipe gap decreased 5 from default 100.\\
Krazy World \cite{Stadie2018} & 10$\times$10$\times$3 & no & Randomly generated worlds from the same distribution.\\
Mazes \cite{Stadie2018} & 20$\times$20 & no & Randomly generated mazes from the same distribution.\\
Arcade Learning Environment \cite{Machado2018} & 210$\times$160$\times$3 & no & 60 different games available in the plaform.\\
CoinRun \cite{Cobbe2018} & 64$\times$64$\times$3 & no & Randomly generated maps with 3 levels of difficulty.\\
CoinRun Platform \cite{Cobbe2018} & 64$\times$64$\times$3 & no & Randomly generated maps from the same distribution.\\
RandomMazes \cite{Cobbe2018} & [3$\times$3 - 25$\times$25] & no & Randomly generated mazes of different sizes.\\
\cmidrule{1-4} 
\textbf{CRLMaze} & \textbf{320$\times$240$\times$3} & \textbf{yes} & Light/visibility, walls textures, object shape and colors are changed within the same object picking task.\\
\bottomrule
\end{tabular}
\end{table*}

For all the scenarios, changes are not gradual but happening at three specific points equidistant in time (the total number of training episodes is fixed and considered a property of the environment) and practically implemented as different ZDoom \emph{maps} faced sequentially ($M_1 \rightarrow M_2 \rightarrow M_3$, see Fig. \ref{img:env-changes}). The agent is randomly spawned at fixed positions depicted as white points in Fig. \ref{img:doom-maze} with a random visual angle. The environment starts with 75 randomly spawned \emph{column} objects and 50 \emph{lantern} objects. Catching a column increases the reward of 100 once collected while touching a lantern decreases the reward of 200. Even if in our exploratory experiments we noted that a shaping reward is not necessary to train the agents up to convergence, a weak shaping reward of 0.7 has been added when the \emph{go-forward} action is chosen to improve environment exploration and ultimately speed-up learning convergence. A new object for each category is also randomly spawned every 3 ticks for roughly maintaining the amount of objects in the environment stable as when the objects are collected by the agent they disappear.

In Tab. \ref{tab:envs}, we report some of the common environments and platforms used in the context of continual and meta-reinforcement learning and compare them with the proposed \emph{CRLMaze}.

While still acknowledging the limited number of environmental variations introduced in the benchmark, we believe CRLMaze shows a significant advancement in terms of task complexity and non-stationary elements introduced with respect to recent environments made available by the community, being them usually of low input dimensionality, not often running on a complex 3D engine and with very limited non-stationary dynamics. For example, in \cite{Riemer2018}, the \emph{Catcher} environment is based on simple 2D physics and the only non-stationary element introduced is a change in the vertical velocity of the pallet which is only slightly increased of 5\% from its default value.   

% \begin{figure}[h]
%   \centering
%   \includegraphics[width=0.60\textwidth]{imgs/doom_objects_and_textures}
%   \caption{\small Objects (top) and Textures (bottom) used for the \emph{CRLMaze} scenarios. Better viewed in colors.}
%   \label{fig:obj_txt}
% \end{figure}

\section{Continual Reinforcement Learning Strategies}
\label{sec:strat}

% Learning in non-stationary envs: meta-learning, hierarchical-rl, continual
% memory consolidation with or without t (no multi-task setting)
% CL4RL (no mem. buffer nor custom parameters)
% Other baselines

Learning over complex and large non-stationary environments is a hard challenge for current reinforcement learning systems. Recent works in this research area include meta-learning \cite{Finn2019a,Nagabandi2018,Al-shedivat2017}, hierarchical learning \cite{Wu2018a,Vezhnevets2016} and continual learning approaches \cite{Kirkpatrick2017,Schwarz2018, Parisi2020}. While both meta-learning and hierarchical learning work around the idea of imposing some structural dependencies among the learned concepts, continual learning is generally agnostic with this regard, being more focused on addressing the non-stationary nature of the underlying distributions \cite{Parisi2019}.

Consolidating and preserving past memories while being able to generalize and learn new concepts and skills is a well known challenge for both artificial and biological learning systems, generally acknowledged as the \emph{plasticity-stability} dilemma \cite{Mermillod2013}. Since gradient-based architectures are generally skewed towards plasticity and prone to \emph{catastrophic forgetting}, much of the research in continual learning with deep architectures has been devoted to the integration of consolidation processes in order to improve stability \cite{Chen2018,Goodfellow2013}. 

However, the general focus of continual reinforcement learning research has been devoted to multi-task scenarios \cite{Kirkpatrick2017,Schwarz2018} where consolidation can be achieved more easily and only when there is a change of task. \emph{CRLMaze} constitutes a step forward in the evaluation of new continual reinforcement learning strategies that have to deal with substantial, unpredictable changes in the environment \emph{within the same task} and without any additional supervised signal indicating (virtual or real) shifts in the underlying input-output distribution. We regard at this situation as the most realistic (and difficult) setting every agent should be able to deal within real-world conditions, where learning is mostly unsupervised and autonomous.

In recent literature, this problem has been tackled by using external generative models of the environment in order to detect big changes in input space \cite{Kirkpatrick2017}. However, recent evidences in behavioral experiments on rats suggests, more generally, behavioral correlates of synaptic consolidation especially when the subject is exposed to novel or strong external stimuli (e.g. a foot shock) \cite{Clopath2012,Clopath2008}. Following this inspiration, in this paper we propose a new strategy \textbf{\emph{CRL-Unsup}}, where the central idea is to consolidate memory only when a substantial difference between the expected reward and the actual one is detected, i.e. when the agent encounters an unexpected situation.

Hence, distributional shifts can be detected just by looking at the ability of the agent to actually perform the task: this can be approximated and practically implemented as the difference between a short-term ($r_{mavg}^s$) and a long-term ($r_{mavg}^l$) reward moving average that, when goes under a particular threshold ($\eta$), triggers the memory consolidation procedure\footnote{It is interesting to note that a similar technique is also the basis of the \emph{MACD indicator} \cite{Seyedi2013} widely used in automated trading systems to detect changing market conditions and issue buy/sell signals.}. The long-term moving average encodes the expected reward over a longer timespan, while the shorter one, an average of the currently received rewards where noise has been partially averaged out. This approach may not only signal changes in the environment affecting the performance of the agent but also possible changes in the reward function or instabilities of the learning process which may be mitigated through consolidation (similar to the regularization loss introduced in PPO \cite{Schulman2017}). Moreover, we do not use neither any distribution-specific over-parametrization nor any kind of memory replay as deemed necessary in \cite{Kirkpatrick2017,Schwarz2018}.

For consolidation in \emph{CRL-Unsup} we employ the end-to-end regularization approach firstly introduced in \cite{Kirkpatrick2017} and known as \emph{Elastic Weight Consolidation} (EWC): the basic idea is to preserve the parameters proportionally to their importance in the approximation of a specific distribution (i.e. the Fisher information). More efficient consolidation techniques through regularization derived from EWC have been recently proposed \cite{Schwarz2018,Zenke2017,maltoni2019,Lomonaco2019}. However, for simplicity, we used its basic implementation. In eq. \ref{eq:ewc_loss} and eq. \ref{eq:lambda} the loss function $L$ of the \emph{CRL-Unsup} strategy for a single consolidation step is reported\footnote{Please note that in the basic EWC implementation a regularization term for each consolidation step needs to be added to the loss with a different $F_k$ and $\theta_k^*$ for each weight $\theta_k$ (see interesting discussion in \cite{Schwarz2018}).} where $L_{A2C}$ is the standard A2C loss function composed of the value and policy loss as defined in \cite{Mnih2016}; $\lambda$ is an hyper-parameter encoding the strength of the consolidation (i.e. reducing plasticity); $F_k$ is the Fisher information for the weight $\theta_k$ while $\theta_k^*$ indicates the optimal weight to consolidate. 
However, since the Fisher information can not be computed on the new data distribution, $F$ is computed at fixed steps in times and only the latest one is used in the regularization term when the threshold is exceeded.

  \begin{equation}
    L = L_{A2C} + \frac{\lambda}{2} \cdot \sum_k F_k (\theta_k - \theta_k^*)^2
    \label{eq:ewc_loss}
  \end{equation}

  \begin{equation}
    \lambda = 
        \begin{cases}
            \alpha &     \quad\text{if } r^s_{mavg} - r^l_{mavg} \le \eta \\
            0 &    \quad\text{otherwise}\\
        \end{cases}
            \label{eq:lambda}
  \end{equation}

\begin{figure}[h]
\begin{center}
\includegraphics[width=0.8\textwidth]{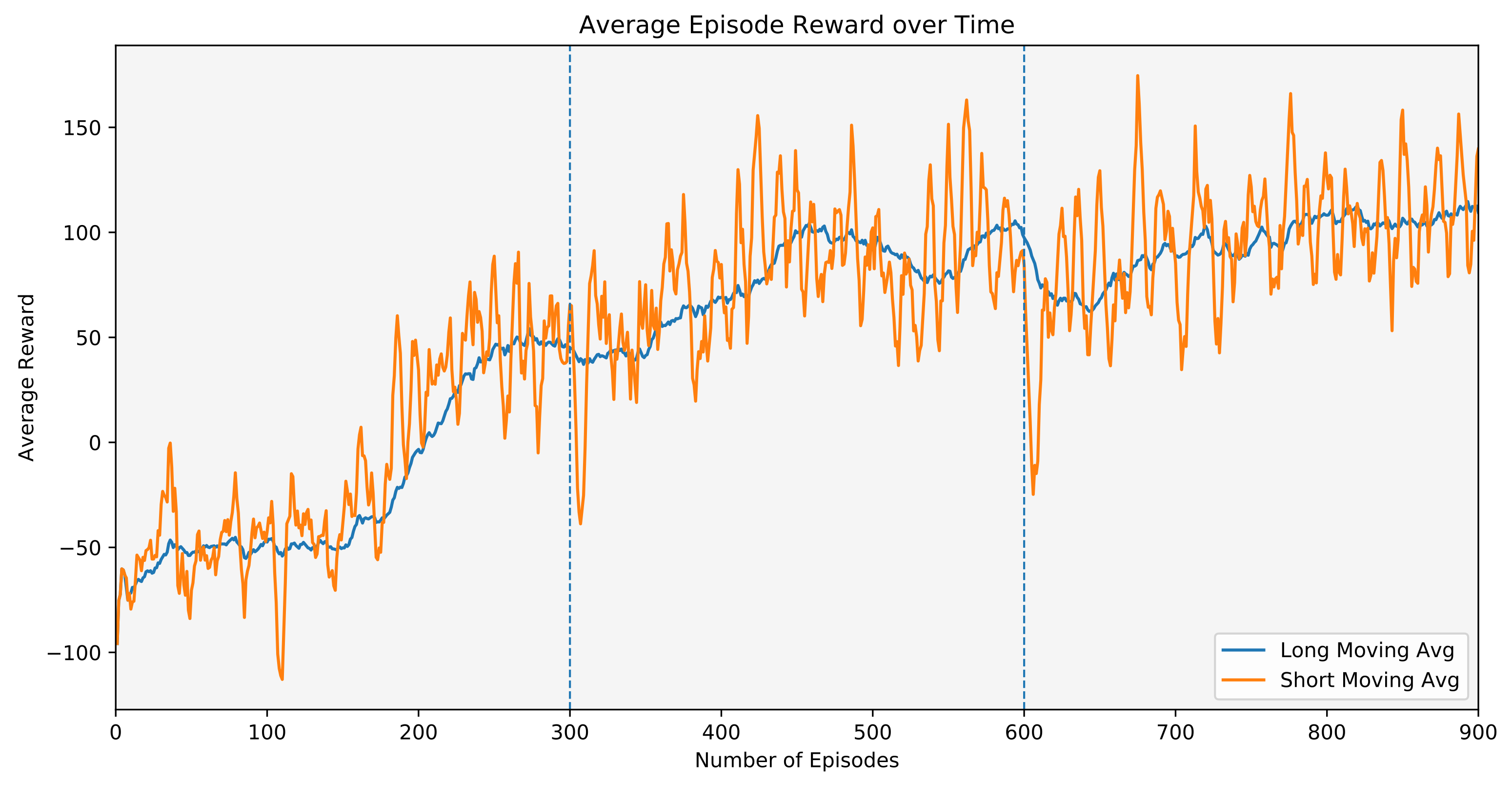}
\includegraphics[width=0.8\textwidth]{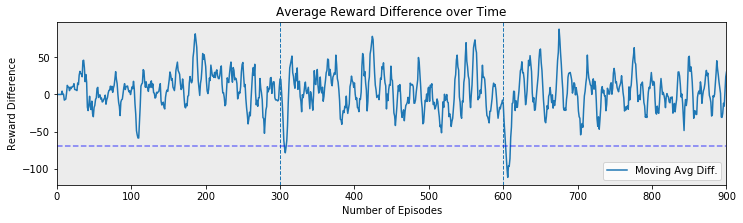}
\caption{\small Short and long-term moving average (computed over 6 and 50 episodes, respectively) of the average cumulative reward during training in the \emph{light} scenario. Dotted lines indicate when the environment is changed. In this example, the difference between the short-term and long-term moving average goes under $\eta = -70$ when the environment changes.}
\label{fig:cl4rl_reward_curve}
\end{center}
\end{figure}

In Figure \ref{fig:cl4rl_reward_curve}, an example in the \emph{light} scenario of the short and long-term moving average (computed over 6 and 50 episodes, respectively) of the training average cumulative reward is reported. In order to better compare and understand the performance of the aforementioned strategy, on each of the considered environmental changes (i.e. \emph{light}, \emph{texture}, \emph{object}, \emph{all}) four different baselines are here introduced and assessed:

\begin{enumerate}

   \item \textbf{\emph{Multienv}}: This approach can be considered as a reference baseline and not properly a CRL strategy since it consists in training the agent over all the possible environmental conditions (i.e. maps $M_1$, $M_2$ and $M_3$) of each scenario at the same time. Having access to all the maps at the same time makes the distribution stationary and eliminates the \emph{catastrophic forgetting} problem. It will be considered as an upper bound for the other strategies as generally acknowledged in continual learning \cite{Parisi2019,pmlr-v78-lomonaco17a,Lopez-paz2017}.

   \item \textbf{\emph{CRL-Naive}}: This approach, like the homonym strategy in the supervised context \cite{pmlr-v78-lomonaco17a,maltoni2019}, consists in just continuing the learning process without variations and indifferently w.r.t. the changes in the environment. Learning through reinforcement in complex non-stationary environments without any \emph{memory replay} is known to suffer from catastrophic forgetting, instability and convergence difficulties while learning. This strategy is usually considered as a lower bound.

   \item \textbf{\emph{CRL-Sup}}: This approach can be considered as a second baseline in which the distributional shift supervised signal (i.e. when the map changes) is actually provided to the model for memory consolidation purposes. In this case the standard application of EWC with the loss described in eq. \ref{eq:ewc_loss} is performed but is perfectly synchronized with the end of the training on each map.

   \item \textbf{\emph{CRL-Static}}: In this strategy, the memory is consolidated (i.e. the regularization term added) at fixed steps in time, independently of the changes in the environment. As we will see in the experiments results, this may be very difficult to tune and rather inefficient, depending on the memory consolidation technique used. In fact, when learning from scratch an early and ``blind'' consolidation of memory may also hurt performance and actually hamper the ability of learning in the future.

\end{enumerate}

\section{Experiments and Results}
\label{sec:exp}

% Metrics and results

% details
For all the experiments we use a simple batched-A2C with synchronous updates \cite{Stooke18}, but only \emph{within} the same \emph{map} (the actual environment with fixed static settings), so that when the map changes the model cannot access in any way previous environmental conditions. The architecture of the agent used for these experiments is a plain 3-layers ConvNet (3$\times$3 kernels with 32 feature maps each) with ReLU activations, followed by a fully connected layer encodings the three possible actions $A$ = \{\emph{turn-left}, \emph{turn-right} and \emph{move-forward}\}\footnote{Input frames with an original resolution of 320$\times$240 are downscaled to 160$\times$120.}. Each training and test episode has a fixed runtime of 1000 ticks. However, the agent is allowed to make an action every 4 frames, maintaining the action chosen based on the first frame fixed for the other three. This allows a smoother interaction with the environment and allows to the agent to not stall in ambiguous situations even if completely stateless (e.g. in front of a wall).  

For the batched-A2C implementation, the synchronous gradient update takes place every 20 frames (covering 80 ticks of the total 1000 ticks of the full episode length) and 20 different agents are spawned in parallel in 20 \emph{ViZDoom} instances of the same environment. The discount factor is fixed to $\gamma = 0.99$ for all the environments. More details about the experimental procedure, implementation details and all the hyper-parameters used are available in the appendix \ref{appendix:exps_details}.

% resume scenarios.
% experiments and results
% tables for each scenario

\begin{table}[!htbp]
%\small
\scriptsize
\centering
 \caption{\small Average cumulative reward matrix $R$ and $A$ metric result for each scenario and CRL strategy. Results highlighted in \textbf{black} and {\color{blue} blue} represent the best and the second best performing strategies on each scenario. A gray background is used in the cells involved in the computation of the $A$ metric. Results are computed over 10 runs for each strategy and benchmark for a total of 160 runs. Standard deviation is reported in Tab. \ref{tab:std_res}}
\label{tab:cl4rl_res}
\setlength\tabcolsep{4pt}
\begin{tabular}{llccc|ccc|ccc|ccc}
\toprule

& & \multicolumn{3}{c}{\textbf{CRL-Naive}} & \multicolumn{3}{c}{\textbf{CRL-Sup}} & \multicolumn{3}{c}{\textbf{CRL-Static}} & \multicolumn{3}{c}{\textbf{CRL-Unsup}}\\
\cmidrule{2-14} 
\parbox[t]{2mm}{\multirow{6}{*}{\rotatebox[origin=c]{90}{Light}}}
& & $M_1$ & $M_2$ & $M_3$ & $M_1$ & $M_2$ & $M_3$ & $M_1$ & $M_2$ & $M_3$ & $M_1$ & $M_2$ & $M_3$ \\
\cmidrule{2-14} 
& $M_1$ &  \cellcolor{gray!25} 93 & -460 & 144 & \cellcolor{gray!25} 283 & -350 & 322 & \cellcolor{gray!25} 242 & -613 & -36 & \cellcolor{gray!25} 334 & -298 & 491\\
& $M_2$ &  \cellcolor{gray!25} -987 & \cellcolor{gray!25} 528 & -842 & \cellcolor{gray!25} -236 & \cellcolor{gray!25} 545 & 642 & \cellcolor{gray!25} -987 & \cellcolor{gray!25} 909 & -551 & \cellcolor{gray!25} -987 & \cellcolor{gray!25} 1090 & -537\\
& $M_3$ & \cellcolor{gray!25} -892 & \cellcolor{gray!25} 1063 & \cellcolor{gray!25} 938 & \cellcolor{gray!25} -232 & \cellcolor{gray!25} 116 & \cellcolor{gray!25} 615 & \cellcolor{gray!25} -800 & \cellcolor{gray!25} 832 & \cellcolor{gray!25} 1106 & \cellcolor{gray!25} -892 & \cellcolor{gray!25} 426 & \cellcolor{gray!25} 818\\
\cmidrule{2-14} 
& \textbf{A} & \multicolumn{3}{c}{123,96} & \multicolumn{3}{c}{\color{blue}181,97} & \multicolumn{3}{c}{\textbf{217,4}} & \multicolumn{3}{c}{131,65}\\
\specialrule{0.8pt}{2pt}{2pt}

\parbox[t]{2mm}{\multirow{6}{*}{\rotatebox[origin=c]{90}{Texture}}}
& & $M_1$ & $M_2$ & $M_3$ & $M_1$ & $M_2$ & $M_3$ & $M_1$ & $M_2$ & $M_3$ & $M_1$ & $M_2$ & $M_3$ \\
\cmidrule{2-14} 
& $M_1$ & \cellcolor{gray!25} 877 & 544 & 57 & \cellcolor{gray!25} 1196 & 1058 & 385 & \cellcolor{gray!25} 1049 & 822 & 152 & \cellcolor{gray!25} 1105 & 836 & 72\\
& $M_2$ & \cellcolor{gray!25} -115 & \cellcolor{gray!25}1360 & 504 & \cellcolor{gray!25}-80 & \cellcolor{gray!25}1415 & 867 & \cellcolor{gray!25}-6 & \cellcolor{gray!25}1150 & 479 & \cellcolor{gray!25}186 & \cellcolor{gray!25}1283 & 631\\
& $M_3$ & \cellcolor{gray!25}-283 & \cellcolor{gray!25}-263 & \cellcolor{gray!25}1422 & \cellcolor{gray!25}-243 & \cellcolor{gray!25}-194 & \cellcolor{gray!25}1352 & \cellcolor{gray!25}-218 & \cellcolor{gray!25}-176 & \cellcolor{gray!25}1121 & \cellcolor{gray!25}-215 & \cellcolor{gray!25}-156 & \cellcolor{gray!25}1252\\
\cmidrule{2-14} 
& \textbf{A} & \multicolumn{3}{c}{499,81} & \multicolumn{3}{c}{\color{blue}574,39} & \multicolumn{3}{c}{486,63} & \multicolumn{3}{c}{\textbf{575,87}}\\
\specialrule{0.8pt}{2pt}{2pt}

\parbox[t]{2mm}{\multirow{6}{*}{\rotatebox[origin=c]{90}{Object}}}
& & $M_1$ & $M_2$ & $M_3$ & $M_1$ & $M_2$ & $M_3$ & $M_1$ & $M_2$ & $M_3$ & $M_1$ & $M_2$ & $M_3$ \\
\cmidrule{2-14} 
& $M_1$ & \cellcolor{gray!25}930 & -974 & -1005 & \cellcolor{gray!25}1365 & -664 & -989 & \cellcolor{gray!25}1129 & -953 & -1006 & \cellcolor{gray!25}1308 & -695 & -995\\
& $M_2$ & \cellcolor{gray!25}962 & \cellcolor{gray!25}1045 & -988 & \cellcolor{gray!25}1160 & \cellcolor{gray!25}1221 & -934 & \cellcolor{gray!25}781 & \cellcolor{gray!25}944 & -937 & \cellcolor{gray!25}992 & \cellcolor{gray!25}1080 & -959\\
& $M_3$ & \cellcolor{gray!25}-758 & \cellcolor{gray!25}-214 & \cellcolor{gray!25}1013 & \cellcolor{gray!25}254 & \cellcolor{gray!25}-125 & \cellcolor{gray!25}878 & \cellcolor{gray!25}-676 & \cellcolor{gray!25}-242 & \cellcolor{gray!25}845 & \cellcolor{gray!25}54 & \cellcolor{gray!25}-131 & \cellcolor{gray!25}840\\
\cmidrule{2-14} 
& \textbf{A} & \multicolumn{3}{c}{496,67} & \multicolumn{3}{c}{\textbf{792,59}} & \multicolumn{3}{c}{463,7} & \multicolumn{3}{c}{\color{blue}690,8}\\
\specialrule{0.8pt}{2pt}{2pt}

\parbox[t]{2mm}{\multirow{6}{*}{\rotatebox[origin=c]{90}{All}}}
& & $M_1$ & $M_2$ & $M_3$ & $M_1$ & $M_2$ & $M_3$ & $M_1$ & $M_2$ & $M_3$ & $M_1$ & $M_2$ & $M_3$ \\
\cmidrule{2-14} 
& $M_1$ & \cellcolor{gray!25}1268 & -1000 & -1000 & \cellcolor{gray!25}1579 & -1000 & -1000 & \cellcolor{gray!25}1132 & -1001 & -1000 & \cellcolor{gray!25}1518 & -998 & -1000\\
& $M_2$ & \cellcolor{gray!25}-490 & \cellcolor{gray!25}1346 & -991 & \cellcolor{gray!25}301 & \cellcolor{gray!25}904 & -999 & \cellcolor{gray!25}-503 & \cellcolor{gray!25}1044 & -999 & \cellcolor{gray!25}-301 & \cellcolor{gray!25}1103 & -1006\\
& $M_3$ & \cellcolor{gray!25}-764 & \cellcolor{gray!25}-219 & \cellcolor{gray!25}815 & \cellcolor{gray!25}-389 & \cellcolor{gray!25}-370 & \cellcolor{gray!25}758 & \cellcolor{gray!25}-496 & \cellcolor{gray!25}-197 & \cellcolor{gray!25}680 & \cellcolor{gray!25}-286 & \cellcolor{gray!25}-332 & \cellcolor{gray!25}695\\
\cmidrule{2-14} 
& \textbf{A} & \multicolumn{3}{c}{325,88} & \multicolumn{3}{c}{\textbf{464,04}} & \multicolumn{3}{c}{276,4} & \multicolumn{3}{c}{\color{blue}399,59}\\
\specialrule{0.8pt}{2pt}{2pt}
\multicolumn{2}{c}{\textbf{Avg. A}} & \multicolumn{3}{c}{361,57 $\pm$ 154,18 } & \multicolumn{3}{c}{\textbf{503,24} $\pm$ 219,96} & \multicolumn{3}{c}{361,03 $\pm$ 116,30} & \multicolumn{3}{c}{{\color{blue}449,47} $\pm$ 210,78} \\
\bottomrule
\end{tabular}
\end{table}

In order to evaluate and compare the performance of each strategy we use the $A$ metric, defined in \cite{Diaz-Rodriguez2018} as an extension of \cite{Lopez-paz2017}. Performance are evaluated at the end of the training on each map $M_i$ on 300 testing episodes, 100 for each different map $M_j$, even the ones not already encountered. Given the test cumulative reward matrix $R \in \mathbb{R}^{3 \times 3}$, which contains in each entry $R_{i,j}$ the \emph{test episodes} average cumulative reward of the model on map $M_j$ after observing the last \emph{training episode} from map $M_i$; $A$ can be defined as follows: 

  \begin{equation}
    A = \frac{\sum_{i \ge j}^{N} R_{i,j}}{\frac{N (N+1)}{2}}
  \end{equation}

where $N=3$ and $A$ is essentially the average of the lower triangular matrix of $R$, which roughly encodes how the model is performing on the current environmental change and the already encountered ones, on average. In Table \ref{tab:cl4rl_res} the average cumulative reward and the $A$ metric results for the \emph{Light}, \emph{Texture}, \emph{Object}, \emph{All} scenarios and the 4 different CRL strategies is reported. It is worth noting that, the scenarios difficulty can vary substantially from a cumulative reward average of $\sim$200 for the agents trained in the \emph{light} scenario to $\sim$600 for the \emph{Object} one, which turns out to be the easiest one in our experiments.

\begin{figure}[h]
  \centering
  \includegraphics[width=0.8\textwidth]{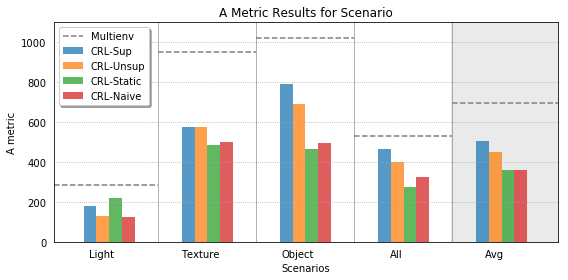}
  \caption{\small $A$ metric results for each CRL strategy and scenario. Dotted lines indicate the average cumulative reward on 300 test episodes for the \emph{multienv} upper-bound. Standard deviation is reported in Tab. \ref{tab:std_multienv} Better viewed in colors.}
  \label{img:hist}
\end{figure}

By considering the average $A$ metric across all the scenarios for each strategy (at the bottom of Table \ref{tab:cl4rl_res}) or in the last column of Fig. \ref{img:hist}, it is possible to compare the different strategies independently of the peculiarities of each specific scenario. In this case we can observe how the \emph{CRL-Sup} strategy constitutes, as we would expect, the best approach in terms of absolute $A$ performance. However, the proposed \emph{CRL-Unsup} strategy, while not exploiting any additional supervised signal, reasonably approximates its performance with a gap of $\sim$50 cumulative reward points. The \emph{CRL-Static} and \emph{CRL-Naive} approaches perform similarly on the $A$ metric, but while the \emph{CRL-Naive} approach is almost consistently better on the last map $M_3$ at the end of the training, it seems more sensitive to forgetting than the \emph{CRL-Static} approach on previously encountered maps.

Results for each specific scenario roughly confirm this trend with exception of the \emph{light} scenario (the most difficult) where a \emph{CRL-Static} approach seems to prevail even the \emph{CRL-Sup} one. We postulate that, in this case, a more frequent consolidation in the direction of the natural gradient as shown in \cite{Schulman2017} may help to stabilize learning in complex environments even within the same environmental conditions.

Finally, in Fig. \ref{img:hist}, the average $A$ metric for each strategy is reported along with the \emph{Multienv} upper bound. In this case the upper bound is not an $A$ metric but simply the average test cumulative reward (on all the test maps) of a agent trained simultaneously on the three environmental conditions of each scenario. It is worth noting the conspicuous gap w.r.t. the best performing continual reinforcement learning strategy of each scenario, suggesting the need of further research on CL approaches for RL.

\section{Discussion and Conclusions}
\label{sec:conclusions}

% consolidation in EWC, biology, etc. [?]
% better CL strat below?
% Integrate prediction on changes in environments (next frame) [x]
% More principles predict your own loss [x]

% lunghi scenario where things can happen again and we cambine mutations randomly
% Consider fixed feature extractor + efficiency improvements
% Combine CL + SRL + RL
% More principled solution embeedding the Unsup solution in the loss function
% Look at the reward but also at the environment apparences with a single model

% conclusions: forward transfer difficult, possible future integration with unsup reconstruction in the same model may help.

% open questions:
% - Positive transfer makes harder to detect change in the reward.
% - What if the change in the environment is gradual?
% - Is consolidation enough for learning continually?

In this work we introduced and openly released a new environment and benchmark for easily assessing continual reinforcement learning algorithms on a complex 3D non-stationary environment. The preliminary experiments introduced in section \ref{sec:exp} on four different scenarios and 5 different strategies show that the proposed unsupervised approach without any distributional shift supervised signal, external model or distribution-specific over-parametrization is not only possible but may be competitive with respect to a standard \emph{supervised} counterpart.

However, as we observed in some experiments (where there is a noticeable gap between the \emph{CRL-Sup} and \emph{CRL-Unsup} strategies), the detection of a timely consolidation signal can be sometimes critical. For example, in case of \emph{positive forward transfer} followed by a possible \emph{negative backward transfer} \cite{Lopez-paz2017} (i.e. being able to perform well on new conditions but impacting negatively on learned knowledge about the previous ones) memory consolidation can not take place just by looking at the training cumulative reward curve since steadily growing. This problem may be tackled by looking at an additional regularization loss for reconstructing the input frame (or predicting the next one) since changes in the input space may be more evident. In this way, while still using a single end-to-end model and constructing more robust features \cite{Hermann2017,TimotheeLesortNataliaDiaz-Rodriguez2018}, it would be possible to integrate the benefit of both approaches when learning continuously.

In the future we plan to expand this work in several other directions. Firstly by moving towards a more flexible and more principled solution where the consolidation is proportional to the expected reward difference encoded directly in the loss function. Secondly by integrating more accurate synaptic plasticity models as shown in \cite{Kaplanis18, benna2016} and going beyond mere consolidation processes which tend to quickly saturate the model learning capacity. 

Finally, we plan to extend our evaluation where existing environmental changes are discretized by providing additional training maps for each category and by adding a new environmental change category where the size of the maze is substantially varied.

While still in their infancy we can foresee a new generation of reinforcement learning algorithms which can learn continually in complex non-stationary environments, opening the door to artificial learning agents which can autonomously acquire new knowledge and skills in unpredictable, real-word settings.

% ---- Bibliography ----

%{\color{red} TODO: fix biblio.}

\medskip
\small

\bibliographystyle{splncs04}
\bibliography{library}

%\clearpage
\appendix
\section{Experiments Details}
\label{appendix:exps_details}

% hardware and training time
% runs for each scenario
% dev.std
% others from checklist?

\normalsize
In this section additional details about the experiments are reported. All the code, environments and setup scripts to re-produce the experiments are openly released at: \url{https://github.com/vlomonaco/crlmaze}. In order to properly compare the performance of the proposed CRL strategies a total of 200 runs (10 for each CRL strategy and scenario) has been conducted for more than 40 hours of computation on a single machine with 32 CPU cores and 1 NVIDIA GTX Titan X%\footnote{Exceptional runs not converging within the predefined number of training episodes have been considered outliers and hence excluded from the results.}.

In Tab. \ref{tab:std_res} and \ref{tab:std_multienv} the standard deviation of the average cumulative reward computed over the testing episodes for each strategy and scenario is reported.

\begin{table}[!htbp]
%\small
\scriptsize
\centering
 \caption{\small Standard deviation of the testing average cumulative reward matrix $R$ presented in Tab \ref{tab:cl4rl_res} and computed over 10 runs for each strategy and scenario.}
\label{tab:std_res}
\setlength\tabcolsep{4pt}
\begin{tabular}{llccc|aaa|ccc|aaa}
\toprule

& & \multicolumn{3}{c}{\textbf{CRL-Naive}} & \multicolumn{3}{c}{\textbf{CRL-Sup}} & \multicolumn{3}{c}{\textbf{CRL-Static}} & \multicolumn{3}{c}{\textbf{CRL-Unsup}}\\
\cmidrule{2-14} 
\parbox[t]{2mm}{\multirow{6}{*}{\rotatebox[origin=c]{90}{Light}}}
& & $M_1$ & $M_2$ & $M_3$ & $M_1$ & $M_2$ & $M_3$ & $M_1$ & $M_2$ & $M_3$ & $M_1$ & $M_2$ & $M_3$ \\
\cmidrule{2-14} 
& $M_1$ & $\pm$224 & $\pm$240 & $\pm$392 & $\pm$126 & $\pm$243 & $\pm$287 & $\pm$176 & $\pm$181 & $\pm$416 & $\pm$166 & $\pm$241 & $\pm$326\\
& $M_2$ & $\pm$3 & $\pm$266 & $\pm$100 & $\pm$431 & $\pm$258 & $\pm$216 & $\pm$7 & $\pm$288 & $\pm$309 & $\pm$8 & $\pm$276 & $\pm$423\\
& $M_3$ & $\pm$103 & $\pm$251 & $\pm$400 & $\pm$348 & $\pm$192 & $\pm$168 & $\pm$132 & $\pm$390 & $\pm$177 & $\pm$138 & $\pm$582 & $\pm$443\\
\specialrule{0.8pt}{2pt}{2pt}

\parbox[t]{2mm}{\multirow{6}{*}{\rotatebox[origin=c]{90}{Texture}}}
& & $M_1$ & $M_2$ & $M_3$ & $M_1$ & $M_2$ & $M_3$ & $M_1$ & $M_2$ & $M_3$ & $M_1$ & $M_2$ & $M_3$ \\
\cmidrule{2-14} 
& $M_1$ & $\pm$271 & $\pm$442 & $\pm$488 & $\pm$392 & $\pm$337 & $\pm$321 & $\pm$272 & $\pm$4 & $\pm$476 & $\pm$181 & $\pm$466 & $\pm$496\\
& $M_2$ & $\pm$63 & $\pm$211 & $\pm$480 & $\pm$260 & $\pm$148 & $\pm$162 & $\pm$269 & $\pm$275 & $\pm$475 & $\pm$500 & $\pm$302 & $\pm$465\\
& $M_3$ & $\pm$21 & $\pm$37 & $\pm$168 & $\pm$66 & $\pm$80 & $\pm$142 & $\pm$225 & $\pm$261 & $\pm$318 & $\pm$128 & $\pm$129 & $\pm$252\\
\specialrule{0.8pt}{2pt}{2pt}

\parbox[t]{2mm}{\multirow{6}{*}{\rotatebox[origin=c]{90}{Object}}}
& & $M_1$ & $M_2$ & $M_3$ & $M_1$ & $M_2$ & $M_3$ & $M_1$ & $M_2$ & $M_3$ & $M_1$ & $M_2$ & $M_3$ \\
\cmidrule{2-14} 
& $M_1$ & $\pm$399 & $\pm$24 & $\pm$9 & $\pm$219 & $\pm$356 & $\pm$38 & $\pm$319 & $\pm$72 & $\pm$14 & $\pm$152 & $\pm$368 & $\pm$38\\
& $M_2$ & $\pm$401 & $\pm$296 & $\pm$26 & $\pm$273 & $\pm$305 & $\pm$72 & $\pm$589 & $\pm$517 & $\pm$102 & $\pm$355 & $\pm$381 & $\pm$37\\
& $M_3$ & $\pm$270 & $\pm$136 & $\pm$419 & $\pm$540 & $\pm$176 & $\pm$284 & $\pm$361 & $\pm$135 & $\pm$171 & $\pm$679 & $\pm$168 & $\pm$290\\
\specialrule{0.8pt}{2pt}{2pt}

\parbox[t]{2mm}{\multirow{6}{*}{\rotatebox[origin=c]{90}{All}}}
& & $M_1$ & $M_2$ & $M_3$ & $M_1$ & $M_2$ & $M_3$ & $M_1$ & $M_2$ & $M_3$ & $M_1$ & $M_2$ & $M_3$ \\
\cmidrule{2-14} 
& $M_1$ & $\pm$156 & $\pm$1 & $\pm$0 & $\pm$202 & $\pm$11 & $\pm$0 & $\pm$334 & $\pm$2 & $\pm$0 & $\pm$135 & $\pm$3 & $\pm$0\\
& $M_2$ & $\pm$558 & $\pm$181 & $\pm$40 & $\pm$570 & $\pm$323 & $\pm$11 & $\pm$514 & $\pm$236 & $\pm$29 & $\pm$427 & $\pm$192 & $\pm$13\\
& $M_3$ & $\pm$267 & $\pm$330 & $\pm$274 & $\pm$320 & $\pm$158 & $\pm$245 & $\pm$357 & $\pm$509 & $\pm$330 & $\pm$307 & $\pm$103 & $\pm$146\\
\bottomrule
\end{tabular}
\end{table}

\begin{table}[!htbp]
%\small
\scriptsize
\centering
 \caption{\small Standard deviation of the testing average cumulative reward computed over 10 runs for the \emph{Multienv} baseline and each scenario.}
\label{tab:std_multienv}
\setlength\tabcolsep{10pt}
\begin{tabular}{caca}

\toprule
\multicolumn{4}{c}{\textbf{Multienv}}\\
\midrule
\emph{Light} & \emph{Texture} & \emph{Object} & \emph{All}\\
\midrule
$\pm$298,71 & $\pm$289,66 & $\pm$213,62 & $\pm$216,07\\
\bottomrule
\end{tabular}
\end{table}

 Hyper-parameters used in the experiments are reported instead in Tab. \ref{tab:hyper}. Hyper-parameters have been chosen for each strategy in order to maximize the $A$ metric at the end of each run. \emph{Parallel instances} indicates the number of ViZDoom instances and agents running in parallel for the roll-outs always fixed to 20. \emph{Episode size} is the number of frames (not considering the skip-rate of 4 as explained in section \ref{sec:exp}) after which a weights update is performed. The $r_{mavg}^{s}$ and $r_{mavg}^{l}$ size parameters represent instead the number of training episodes to consider for the short and long-term moving average, respectively. 

 Focusing only on the strategies employing consolidation $\eta$, $\alpha$ and $\lambda$ are the parameters already described in section \ref{sec:exp} while \emph{Fisher freq.}, \emph{Fisher clip} and \emph{Fisher sample size} represent respectively \emph{i)} the computing frequency of the fisher matrix in terms of training episodes, \emph{ii)} the clipping value of the importance magnitude as described in \cite{maltoni2019}, and \emph{iii)} the number of episodes used to estimate the fisher information of each weight. 

 Please note that in Tab. \ref{tab:hyper}, the number of \emph{Train episodes} and \emph{Test episodes} is intended as for each stationary condition (i.e. maps for the CRL strategies).

\begin{table}[!htbp]
%\small
\scriptsize
\centering
 \caption{\small Specific hyper-parameters used for each strategy and scenario.}
\label{tab:hyper}
\setlength\tabcolsep{4pt}
\begin{tabular}{llc|a|c|a|c}
\toprule

& & \textbf{CRL-Naive} & \textbf{CRL-Sup} & \textbf{CRL-Static} & \textbf{CRL-Unsup}& \textbf{Multienv}\\
\cmidrule{2-7} 
\parbox[t]{2mm}{\multirow{14}{*}{\rotatebox[origin=c]{90}{Light}}}
& Parallel instances & 20 & 20 & 20 & 20 & 20\\
& Learning rate & 6e-5 & 9e-5 & 9e-5 & 9e-5 & 6e-5\\
& Discount factor & 0.99 & 0.99 & 0.99 & 0.99 & 0.99\\
& Episode Size & 20 & 20 & 20 & 20 & 20 \\
& Train episodes & 300 & 300 & 300 & 300 & 600 \\
& Test episodes & 100 & 100 & 100 & 100 & 100 \\
& $r_{mavg}^l$ size & n.d. & n.d. & n.d. & 50 & n.d.\\
& $r_{mavg}^s$ size & n.d. & n.d. & n.d. & 6 & n.d.\\
& $\eta$ & n.d. & n.d. & n.d. & -80 & n.d.\\
& $\alpha$ & n.d. & n.d. & n.d. & 10e7 & n.d. \\
& $\lambda$ & n.d. & 10e7 & 10e5 & n.d. & n.d. \\
& Fisher freq. & n.d. & 300 & 100 & 100 & n.d. \\
& Fisher clip & n.d & 10e-7 & 10e-7 & 10e-7 & n.d.\\
& Fisher sample size & n.d & 100 & 100 & 100 & n.d. \\
\specialrule{0.8pt}{2pt}{2pt}

\parbox[t]{2mm}{\multirow{14}{*}{\rotatebox[origin=c]{90}{Texture}}}
& Parallel Instances & 20 & 20 & 20 & 20 & 20\\
& Learning rate & 9e-5 & 2e-4 & 2e-4 & 2e-4 & 9e-5\\
& Discount factor & 0.99 & 0.99 & 0.99 & 0.99 & 0.99\\
& Episode Size & 20 & 20 & 20 & 20 & 20 \\
& Train episodes & 300 & 300 & 300 & 300 & 600 \\
& Test episodes & 100 & 100 & 100 & 100 & 100 \\
& $r_{mavg}^l$ size & n.d. & n.d. & n.d. & 50 & n.d.\\
& $r_{mavg}^s$ size & n.d. & n.d. & n.d. & 6 & n.d.\\
& $\eta$ & n.d. & n.d. & n.d. & -50 & n.d.\\
& $\alpha$ & n.d. & n.d. & n.d. & 5e6 & n.d. \\
& $\lambda$ & n.d. & 5e6 & 5e6 & n.d. & n.d. \\
& Fisher freq. & n.d. & 300 & 100 & 100 & n.d. \\
& Fisher clip & n.d & 10e-7 & 10e-7 & 10e-7 & n.d.\\
& Fisher sample size & n.d & 60 & 60 & 60 & n.d. \\
\specialrule{0.8pt}{2pt}{2pt}

\parbox[t]{2mm}{\multirow{14}{*}{\rotatebox[origin=c]{90}{Object}}}
& Parallel Instances & 20 & 20 & 20 & 20 & 20\\
& Learning rate & 9e-5 & 2e-4 & 2e-4 & 2e-4 & 2e-4\\
& Discount factor & 0.99 & 0.99 & 0.99 & 0.99 & 0.99\\
& Episode Size & 20 & 20 & 20 & 20 & 20 \\
& Train episodes & 500 & 500 & 500 & 500 & 2600 \\
& Test episodes & 100 & 100 & 100 & 100 & 100 \\
& $r_{mavg}^l$ size & n.d. & n.d. & n.d. & 50 & n.d.\\
& $r_{mavg}^s$ size & n.d. & n.d. & n.d. & 6 & n.d.\\
& $\eta$ & n.d. & n.d. & n.d. & -60 & n.d.\\
& $\alpha$ & n.d. & n.d. & n.d. & 3e6 & n.d. \\
& $\lambda$ & n.d. & 3e6 & 3e6 & n.d. & n.d. \\
& Fisher freq. & n.d. & 500 & 100 & 100 & n.d. \\
& Fisher clip & n.d & 10e-7 & 10e-7 & 10e-7 & n.d.\\
& Fisher sample size & n.d & 60 & 60 & 60 & n.d. \\
\specialrule{0.8pt}{2pt}{2pt}

\parbox[t]{2mm}{\multirow{14}{*}{\rotatebox[origin=c]{90}{All}}}
& Parallel Instances & 20 & 20 & 20 & 20 & 20\\
& Learning rate & 9e-5 & 2e-4 & 2e-4 & 2e-4 & 2e-4\\
& Discount factor & 0.99 & 0.99 & 0.99 & 0.99 & 0.99\\
& Episode Size & 40 & 40 & 40 & 40 & 40 \\
& Train episodes & 500 & 500 & 500 & 500 & 2600 \\
& Test episodes & 100 & 100 & 100 & 100 & 100 \\
& $r_{mavg}^l$ size & n.d. & n.d. & n.d. & 50 & n.d.\\
& $r_{mavg}^s$ size & n.d. & n.d. & n.d. & 6 & n.d.\\
& $\eta$ & n.d. & n.d. & n.d. & -100 & n.d.\\
& $\alpha$ & n.d. & n.d. & n.d. & 1e6 & n.d. \\
& $\lambda$ & n.d. & 7e6 & 3e6 & n.d. & n.d. \\
& Fisher freq. & n.d. & 500 & 166 & 166 & n.d. \\
& Fisher clip & n.d & 10e-7 & 10e-7 & 10e-7 & n.d.\\
& Fisher sample size & n.d & 60 & 60 & 60 & n.d. \\
\bottomrule
\end{tabular}
\end{table}

\end{document}